\documentclass{llncs}

\usepackage[utf8]{inputenc}
\usepackage{amsmath,amssymb}
\usepackage{graphicx}
\usepackage{url}
\usepackage[disable]{todonotes}
\usepackage{comment}

\begin{document}

\title{Self-Wiring Question Answering Systems}
\author{
Ricardo Usbeck\inst{1} \and 
Jonathan Huthmann\inst{1}  \and
Nico Duldhardt\inst{1}  \and
Axel-Cyrille Ngonga Ngomo\inst{1}}
\institute{AKSW Group, University of Leipzig, Germany\\
\email{usbeck|ngonga@informatik.uni-leipzig.de}}


\maketitle
\begin{abstract}
Question answering (QA) has been the subject of a resurgence over the past years.
The said resurgence has led to a multitude of question answering (QA) systems being developed both by companies and research facilities.
While a few components of QA systems get reused across implementations, most systems do not leverage the full potential of component reuse.  
Hence, the development of QA systems is currently still a tedious and time-consuming process. 
We address the challenge of accelerating the creation of novel or tailored QA systems by presenting a concept for a self-wiring approach to composing QA systems. 
Our approach will allow the reuse of existing, web-based QA systems or modules while developing new QA platforms.
To this end, it will rely on QA modules being described using the Web Ontology Language.
Based on these descriptions, our approach will be able to automatically compose QA systems using a data-driven approach automatically. 
\end{abstract}

\keywords{Question Answering, Content Representation and Processing, System composition, Work in Progress}

\section{Introduction}

\begin{figure}[htb!]
    \centering
    \includegraphics[width=\linewidth]{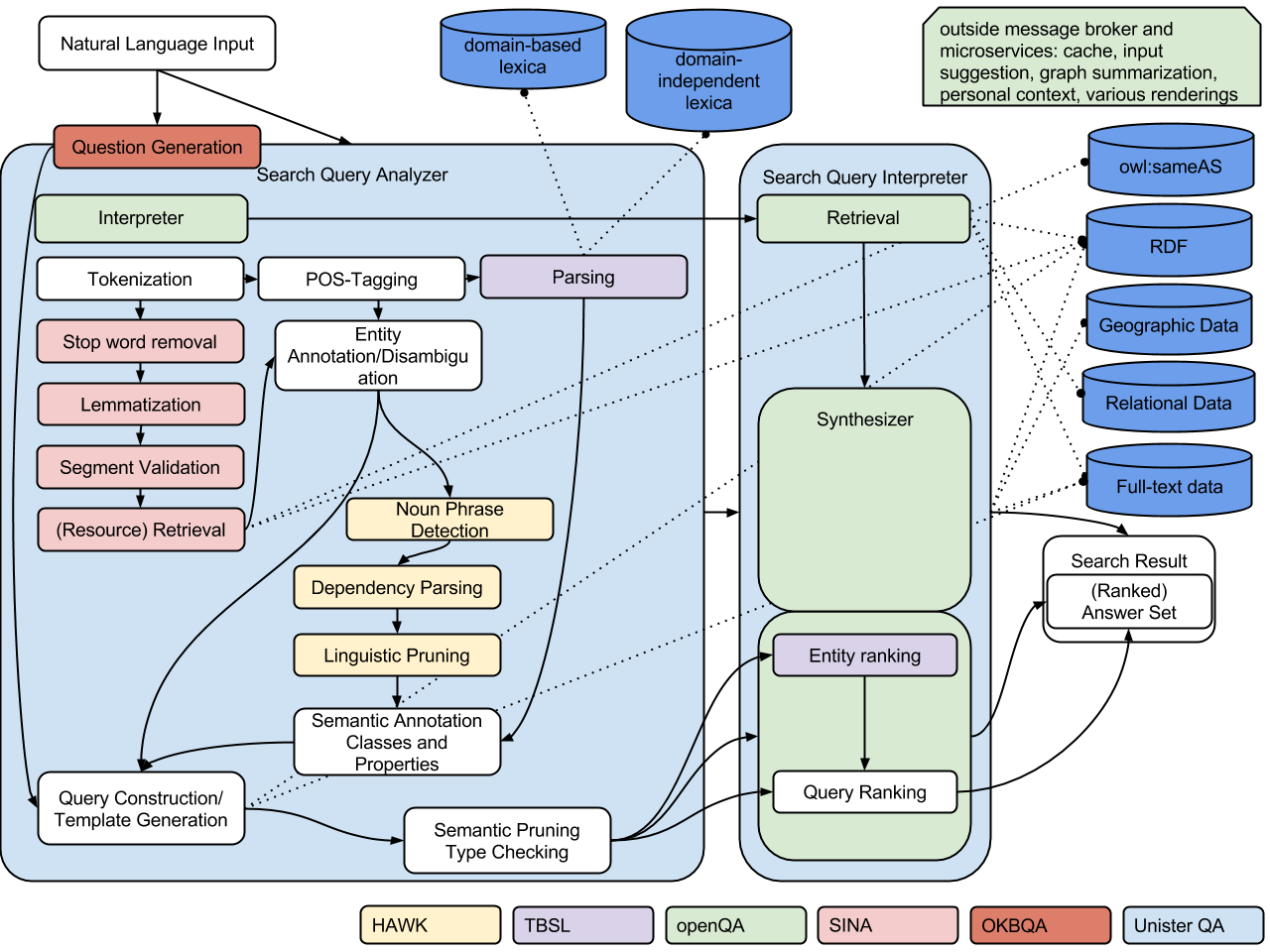}
    \caption{Modules of well-known systems and their possible interactions.}
    \label{fig:overviewOfexistingApproaches}
\end{figure}

More than 20 QA approaches have been solely proposed to the QALD challenge series~\cite{qald6} over the past 6 years.
Most approaches rely on basic modules such as Named Entity Recognizers and linkers, POS-Taggers, SPARQL-executers or graph-traversal modules~\cite{qasurvey}. 
However, most research or industry teams need to start from scratch with implementing a QA system due to (1) a missing awareness of existing modules or (2) existing modules being hard to reuse.
Figure~\ref{fig:overviewOfexistingApproaches} illustrates the problem tackled by this work. It shows dependencies between the components of 6 QA systems~\cite{SINA_WebSemantic,tbsl,hawk,openQA,both2014service}. These dependencies could be used to create and evaluate 20 different QA systems, of which some could potentially perform better than the 6 original systems.

We propose a novel data-driven approach that enables the automatic generation of QA systems.
Our approach assumes the existence of a central registry of QA-related modules. 
Each module in the library is either (1) available as a web-based service which either outputs some form of RDF or (2) such that its output can be mapped via our registry to a well-known RDF vocabulary. 
Given a set of modules in the registry, we try to find links between modules using inference over input and output behaviour of modules.
Afterwards, we use an approach based on refinement operators to wire several modules together to a QA system.
Our contributions are as follows:
\begin{enumerate}
\item We present the first approach for the automatic construction (i.e., wiring) of QA systems from a \textbf{format-agnostic webservice registry}. This registry allows to register already existing modules independent of their input and output format as long as it can be mapped to RDF. 
\item Our approach is \textbf{self-wiring}, i.e., the sequence of service modules a question passes is determined by an automatic discovery algorithm based on the required information of the next module. The required information are based on the input and output behaviour of the registered modules which is already in RDF or was mapped to RDF. Possible links are found using OWL class expressions.
\item We \textbf{analyse existing systems} to find out which modules can be reused and started mapping them into our registry. Furthermore, we are working on a first prototype of the system in our project repository.
\end{enumerate}

The results of this work will enable developers and researchers to focus on the implementation or improvement of single sub-task modules instead of whole systems. 
This way, our approach enables the data-driven vocabulary-independent creation of systems tailored to perform well in the use case for which they are foreseen.
More information can be found at our repository \url{http://OmittedForReview}.


\section{Related Work}
With the growing amount of published QA systems also the search for an universal framework for reusing components began.

One of the earliest works is openQA~\cite{openQA} which is a modular open-source framework for implementing QA systems. 
openQA's main work-flow consists of four stages ({interpretation}, {retrieval}, {synthesis} and {rendering}) as well as adjacent modules ({context} and {service}) written as rigid Java interfaces.
The authors claim that {openQA} enables a conciliation of different architectures and methods.

QALL-ME~\cite{qallme} is another open source approach using an architecture skeleton for multilingual QA, a domain- as well as a domain-independent ontology.
The underling SOA architecture features several web service which are composed to a QA system in a predetermined way.

The open source system OAQA~\cite{oaqa} aims at advancing the engineering of QA systems by following architectural commitments to components for interchangeability.
Using these shared interchangeable components OAQA is able to search the most efficient combination of modules for a task at hand.

QANUS~\cite{qanus} is a not disclosed QA framework for the rapid development of novel QA systems as well as a baseline system for benchmarking.
It was designed to have interchangeable components in a pre-seeded system and comes with a set of common modules such as named entity recognition or part-of-speech tagging.

Both et al.~\cite{both2014service} described a first semantic approach towards coupling components together via RDF to tailor search pipelines using semantic, geospatial and full text search modules.
Here, modules add semantic information to a query until the search intend can be solved. 

QANARY~\cite{singhqanary} is the first real implementation of a semantic approach towards generating QA systems from components. 
Using the provided QA ontology from QANARY, modules can be exchanged, e.g. various versions of NER tools, to benchmark various pipelines and choose the best performing one.

The feasibility of our approach is supported by works from other domains. For example, Verborgh et al.~\cite{DBLP:journals/corr/VerborghAHRMSG15} developed RestDesc which allows for the automatic composition of HyperMedia APIs driven by RDF and Reasoning over N3. 
However, we are not aware of any QA web-modules which are either HyperMedia APIs or are described in a way directly usable in the proposed framework.

\section{Approach}
\subsection{Preliminaries}
In the following, we describe our self-wiring QA architecture formally. 
We call $K = \{(s,p,o)| s \in (U \cup B), p \in U, o \in (U \cup B \cup L) \}$ an RDF knowledge base when $U$ is the set of all Internationalized Resource Identifiers (IRIs), $B$ is the set of all blank nodes and $L$ is the set of all literals following the RDF algebra.\footnote{\url{http://www.w3.org/TR/rdf11-concepts/}}
Our self-wiring architecture consists mainly of two parts: messages $\mu \in M$ and modules $\phi \in \Phi$. 

We call any input or output of a module a \emph{message}.  
Each message $\mu$ contains RDF describing information about the input question and must abide by an OWL class expression $C$ from a knowledge base $K$:
\begin{equation}
\mu = (C), C \in K, \mu \in M
\end{equation}

Messages are handled by modules $\phi \in \Phi$:

\begin{equation}
\phi = (\textrm{input},\mu ,\textrm{output}, \textrm{url}, \textrm{map}),
\end{equation}

where $\textrm{input}$ is an OWL class expression\footnote{\url{http://www.w3.org/TR/owl2-syntax/#Introduction}} used for checking the validity of action on the message $\mu$, see Section~\ref{sec:systemconstruction}.
$\textrm{map}$ is a mapping of URL parameters to literal values of the RDF-based message $\mu$.
\textrm{url} is the URL of the respective webservice and 

\begin{equation}
\textrm{output} = \mu + \delta, \delta \in K.
\end{equation}

\todo[inline]{Maybe $MS(\phi)$ or $\mathbb{N}^{\phi}$}
A system $\Pi$ is defined as 
\begin{equation}
\Pi = I(2^{\Phi}).
\end{equation}
That is, each pipeline consists of an instantiation $I$ of a multi-set of modules. 


\subsection{System Construction}\label{sec:systemconstruction}
By using the formal definition, each generated pipeline starts with a module $\phi_{start}$ with
\begin{equation}
\phi_{start} = (q,\emptyset,\mu_0,\textrm{URI}_i,\textrm{map}_i),
\end{equation}

where $\mu_0$ is an initial message, i.e., a question. Formally not having an input means that one can be connected with everything given that $\emptyset \sqsubseteq C$, where $C$ is any OWL class. Thus let $q$ be our formal input for start modules.

To find out whether the output of module $\phi_i$ is usable for module $\phi_j$ our system evaluates the OWL class expressions that describe the input and output of the modules $\phi_i$ and $\phi_j$. It basically checks where whether $\mbox{output}(\phi_i) \sqsubseteq \mbox{input}(\phi_j)$,
i.e., whether $\phi_j$ would be able to deal with the output of $\phi_i$.
Given this simple operator that allows connecting two modules, we can now define a refinement operator which aims to find \emph{complete} pipelines, i.e., sequences of modules with the \texttt{question} type as input and the \texttt{answer} type as output. The idea behind the refinement is to start with a refinement directed acyclic graph (DAG) which contains only input modules (i.e., modules with input type = \texttt{question}). In each iteration, the refinement operator expands the leaf in the 
DAG that (1) does not yet stands for a sequence of modules with output type = \texttt{answer} and (2) achieves the highest score, where the score function combine the distance between the type of current output and \texttt{answer} in the ontology to drive the platform. Note that this approach is complete, ergo, it is ensured to find all possible pipelines.


Output modules are do not have any further output and contain an RDF resultset in $\mu_\omega$. Again, formally not having an output means that one can be connected with everything given that $\emptyset \sqsubseteq C$ and thus let $a$ be our final output.

\begin{equation}
\phi_{end} = (\textrm{input},a,\mu_\omega,\textrm{URI}_j,\textrm{map}_j))
\end{equation}

\subsection{Supported Vocabularies}

Currently, our extensible, self-wiring system uses 5 ontologies mainly used for Natural Language Programming.
First, there is NIF, the Natural Language Programming Interchange Language which is commonly by Named Entity Recognition and Linking tools~\cite{gerbil} and second Open Annotation, a W3C initiative to annotate any web content\footnote{\url{https://www.w3.org/TR/annotation-vocab/}}.
Furthermore, we are able to use the QANARY ontology which is a relatively new vocabulary designed for QA.
Finally, we will make use of Framenet~\cite{framenet} due to its capability to annotate relations as well as the Earmark ontology\footnote{\url{http://www.essepuntato.it/lode/imported/http://www.essepuntato.it/2008/12/earmark}} used in the state-of-the-art machine reader FRED~\cite{FRED}.


\section{Conclusion}
We introduced a work-in-progress system which able to self-wire new question answering systems from a registry of modules.
Our approach is able to detect all possible QA pipelines and thus enable researchers and developers to focus on particular modules for improvement or enhancement.

Currently, many modules which already use an ontology such as NIF or Open Annotation which could be wired according to our analysis cannot be linked because the underlying ontologies are lacking matching OWL expressions.
Thus, our next step is to run a linking framework over these ontologies to be able to link more modules.

In the future, our approach will leverage{provenance-based} pipeline construction. That is, service modules are adding information to messages which leaves provenance data on the processed message. Thus, users can prevent or force the creation of a pipeline using a particular processing step.
Furthermore, it will be self-improving, i.e., slow services with high response time while multi-user requests 
can be identified while composition and thus benchmarking modules will be able to start particular modules 
more often to improve response time and alert developers of these cases. Thus, the generated pipelines can be tuned towards an {user-given} benchmark dataset.

With this work, we hope to get feedback and establish an active community around an open registry for QA modules.

\bibliographystyle{abbrv}
\bibliography{bib.bib}

\end{document}